\documentclass{article}
\usepackage{ijcai22}

\usepackage{times}
\usepackage{soul}
\usepackage{url}
\usepackage[hidelinks]{hyperref}
\usepackage[utf8]{inputenc}
\usepackage[small]{caption}
\usepackage{graphicx}
\usepackage{amsmath}
\usepackage{amsthm}
\usepackage{amssymb}
\usepackage{dsfont}
\usepackage{booktabs}
\usepackage{algorithm}
\usepackage{algorithmic}
\usepackage{multirow}
\usepackage{subfigure}
\usepackage{color}

\usepackage{array}

\title{Propose-and-Refine: A Two-Stage Set Prediction Network for \\Nested Named Entity Recognition}

\author{
Shuhui Wu\and
Yongliang Shen\and
Zeqi Tan\And
Weiming Lu\thanks{Corresponding author}
\affiliations
College of Computer Science and Technology, Zhejiang University\\
\emails
\{shwu, syl, zqtan, luwm\}@zju.edu.cn
}

\begin{document}

\maketitle

\begin{abstract}
 Nested named entity recognition (nested NER) is a fundamental task in natural language processing. Various span-based methods have been proposed to detect nested entities with span representations. However, span-based methods do not consider the relationship between a span and other entities or phrases, which is helpful in the NER task. Besides, span-based methods have trouble predicting long entities due to limited span enumeration length. To mitigate these issues, we present the Propose-and-Refine Network (PnRNet), a two-stage set prediction network for nested NER. In the propose stage, we use a span-based predictor to generate some coarse entity predictions as entity proposals. In the refine stage, proposals interact with each other, and richer contextual information is incorporated into the proposal representations. The refined proposal representations are used to re-predict entity boundaries and classes. In this way, errors in coarse proposals can be eliminated, and the boundary prediction is no longer constrained by the span enumeration length limitation. Additionally, we build multi-scale sentence representations, which better model the hierarchical structure of sentences and provide richer contextual information than token-level representations. Experiments show that PnRNet achieves state-of-the-art performance on four nested NER datasets and one flat NER dataset.
\end{abstract}

\section{Introduction}

Named Entity Recognition (NER) aims to detect the span and category of all entities in text, which is an essential task in natural language processing. Notably, named entities are often nested in other external entities. For instance, in the sentence ``This indeed was one of Uday's homes", the entity ``Uday" is nested in the entity ``Uday's homes" while ``Uday's homes" is also nested in another larger entity ``one of Uday's homes". This is because natural language sentences are hierarchical. Smaller-scale entities might be nested in larger-scale entities as sub-constituency trees.


\begin{figure}[tb]
  \centering
  \includegraphics[width=\linewidth]{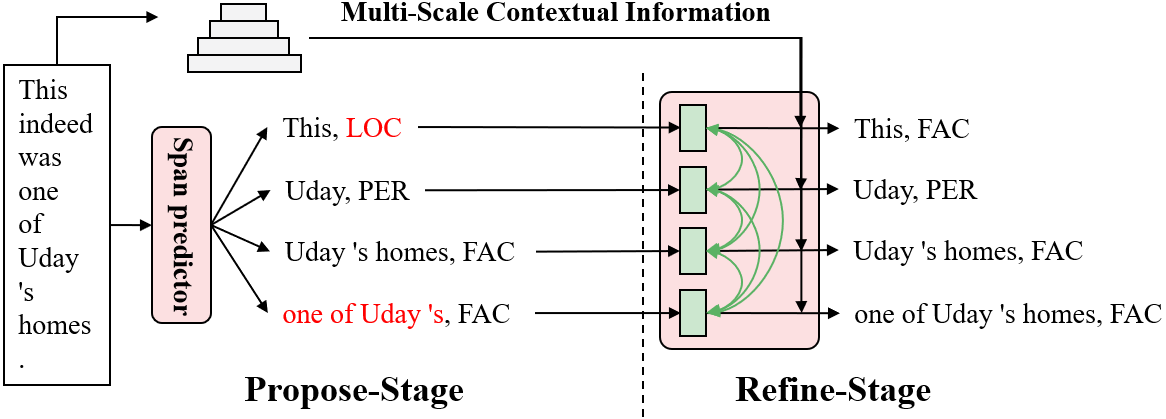}
  \caption{Span-based predictors are error-prone (we color prediction errors in red). The entity ``This" is misclassified due to a lack of interaction with other related phrases in span-based predictors. Besides, Span-based methods cannot predict long entities ``one of Uday 's homes" if we set a small enumeration length limit. PnRNet addresses these issues with proposal refinement and re-prediction. }
  \label{fig:TwoStageIntro}
\end{figure} 

Various methods have been proposed to handle the nested NER task, such as optimized sequence-tagging methods \cite{ju2018neural,strakova2019neural}, hypergraph methods \cite{lu2015joint,katiyar2018nested}, transition-based methods \cite{wang2018neural}. These methods, however, either require complex manual-designed tagging schemas or suffer from error propagation. Recently, span-based methods, which perform classification over features of candidate spans, have gained popularity and have achieved promising results in the NER task \cite{sohrab2018deep,tan2020boundary,shen2021locate,wang2020pyramid}. Unlike previous methods, span-based prediction can naturally address the nested NER task without complex detecting schemas and does not suffer from error propagation. However, as shown in Figure \ref{fig:TwoStageIntro}, span-based methods still have the following two issues. First, the prediction of an entity may rely on other phrases in the sentence. But span representations are typically generated through features of tokens that constitute the span. Therefore the relationship between a span and other phrases or entities is not considered in span-based methods, making the span-based methods error-prone. Second, the length of the enumerated span is always limited since exhaustive enumeration is computationally expensive. Therefore it is hard for span-based methods to detect long entities. 

This paper presents the Propose-and-Refine Network (PnRNet), a two-stage set prediction network for the nested NER. To address the two previously mentioned issues of the span-based methods, we apply a two-stage decoding procedure to detect named entities, as shown in Figure \ref{fig:TwoStageIntro}. In the propose stage, we use a span-based predictor to generate a set of coarse entity predictions as proposals. In the refine stage, proposals are fed into the transformer decoder \cite{vaswani2017attention}, where proposals interact with each other, and richer contextual information is aggregated into the proposal representations. Finally, the refined proposal representations are used to re-predict entity boundaries and classes. In this way, the prediction errors of the coarse proposals can be eliminated with enriched information, and the boundary prediction is not constrained by the enumeration length limitation of the span-based predictor. The final predictions are considered as a set, and a permutation-invariant loss is applied to train the model.

Additionally, we build multi-scale sentence representations to provide richer contextual information in the decoder. As mentioned before, natural language sentences are hierarchical. Therefore, representing the input sentence as a hierarchical structure is natural and helps solve the nested NER problem. For that purpose, we collect the span features generated in the propose stage to form multi-scale sentence representations. In this way, proposals can directly interact with features of spans highly related to the predicted entity rather than token features in the refine stage, which can aggregate hierarchical contextual information more effectively.

Our main contributions are as follows:

\begin{itemize}
    \item We present a novel propose-and-refine two-stage set prediction network for the nested NER task. With richer contextualized information aggregated in the refine stage, PnRNet can make more precise predictions than the span-based predictor. Moreover, PnRNet is not constrained by the span enumeration length because we re-predict entity boundaries and classes after proposal refinement.
    \item To model the hierarchical structure of natural language sentences and better detect nested named entities, we build multi-scale features for decoding to provide richer hierarchical contextual information. 
    \item Experiments on ACE04, ACE05, GENIA, KBP17, and CoNLL03 show that our model outperforms all previous models. We also conduct a detailed ablation study to validate the effectiveness of these innovations.
\end{itemize}

\section{Model}

\begin{figure*}[ht]
    \centering
    \includegraphics[width=\textwidth]{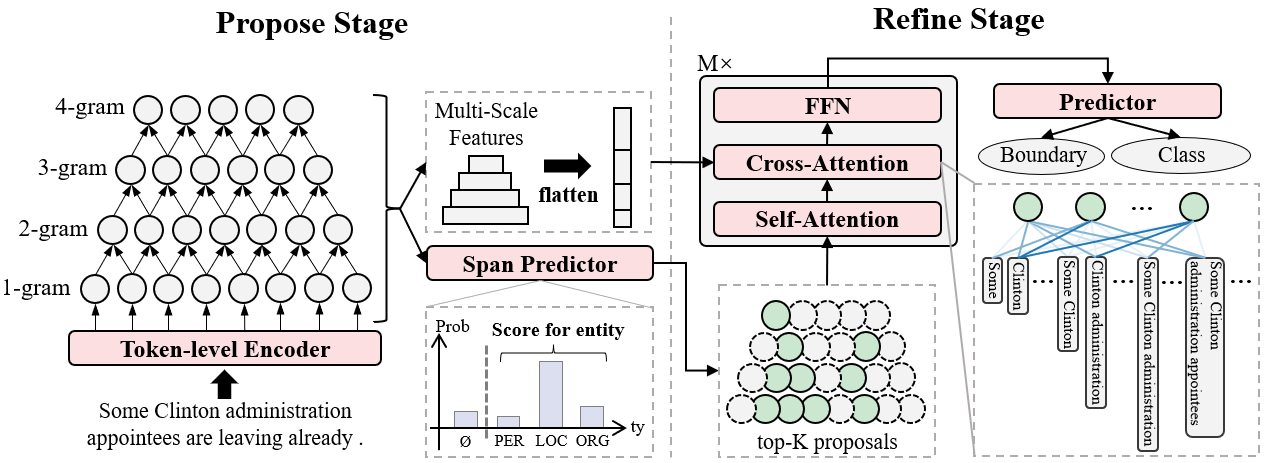}
    \caption{Overview of PnRNet. In the propose stage, PnRNet computes span representations and generates coarse entity proposals with a span-based predictor. In the refine stage, the proposals are refined through a transformer decoder and finally are used to re-predict boundaries and entity classes. We collect multi-scale features from span features generated in the propose stage to provide hierarchical contextual information in proposal refinement. For simplicity of demonstration, we show a PnRNet with span enumeration length limited to $L=4$.}
    \label{fig:ModelOverview}
\end{figure*}

In this section, we will introduce PnRNet, a two-stage set prediction network for nested NER, as illustrated in Figure \ref{fig:ModelOverview}. 


\subsection{Stage \uppercase\expandafter{\romannumeral 1}: Propose} 
\label{sec:Stage1}

\paragraph{Span Feature Generation.} Given an input sentence $\mathbf{X}$ of length $N$, we concatenate the contextual embedding $\mathbf{x}_i^{\text{plm}}$ generated by a pre-trained language model, word embedding $\mathbf{x}_i^{\text{word}}$, part-of-speech embedding $\mathbf{x}_i^{\text{pos}}$, and character embedding $\mathbf{x}_i^{\text{ch}}$ of each token, and then feed the concatenated embeddings into a BiLSTM \cite{hochreiter1997long} for token-level representation $\mathbf{x}_i$:

\begin{equation}
    \label{eq:TokenRepr}
    \mathbf{x}_i=\operatorname{BiLSTM}([\mathbf{x}_i^{\text{plm}}; \mathbf{x}_i^{\text{word}}; \mathbf{x}_i^{\text{pos}}; \mathbf{x}_i^{\text{ch}}])
\end{equation}

\noindent where $[;]$ denotes concatenation operation.

We generate span features from token-level sentence representations in a bottom-up manner:

\begin{equation}
    \mathbf{h}_{l,i}=\begin{cases}
        \operatorname{Linear}([\mathbf{h}_{l-1,i}; \mathbf{h}_{l-1,i+1}])  & \text{ if } l>  1 \\
        \mathbf{x}_i  & \text{ if } l=1
    \end{cases}
\end{equation}

\noindent where $\mathbf{h}_{l,i}$ denotes the feature of the span $(l,i)$, which is the $l$-gram span starting from the $i$-th token. We limit the bottom-up construction process to spans of length $L$ since exhaustive span enumeration is computationally expensive, especially for long sentences.

\paragraph{Entity Proposal.} A span-based predictor is used to classify the entity type of each span with the span features generated in the previous step. The classification scores of span $(l,i)$ is computed as follows:

\begin{equation}
    \mathbf{p}^{\text{cls}}_{l,i}=\operatorname{Softmax}(\operatorname{Linear}(\mathbf{h}_{l,i}))
\end{equation} 

Then the likelihood of that span being an entity can be obtained by:

\begin{equation}
    p_{(l,i)\in \mathcal{E}}=\sum_{\text{ty}_t\ne \varnothing}\mathbf{p}^{\text{cls}}_{l,i}(\text{ty}_t)
\end{equation}

\noindent where $\mathbf{p}^{\text{cls}}_{l,i}(\text{ty}_t)$ indicates the probability of the span $(l,i)$ to be an entity of type $\text{ty}_t$. $\mathcal{E}$ represents all entities in the sentence and $\varnothing$ is a pseudo entity type which means this span is not an entity. 

Span features are sorted by $p_{(l,i)\in \mathcal{E}}$ in descending order, and top-K span features which are most likely to be entities will be picked as the entity proposals $\mathbf{Q}\in \mathds{R}^{K\times d}$. 

It is worth noting that in the nested NER task, the prediction of an entity may rely on other related phrases or entities. However, the span-based predictor does not model the relationship between a span and other phrases. Therefore, the span-based predictor is error-prone, and these entity proposals are just coarse predictions. We have to incorporate richer contextual information into the proposal representation in the refine stage to get more precise predictions.

\subsection{Stage \uppercase\expandafter{\romannumeral2}: Refine}
\label{sec:Stage2}

PnRNet uses a transformer decoder \cite{vaswani2017attention} to refine the coarse entity proposals. The transformer decoder is composed of a stack of $M$ transformer decoder layers. We denote $\mathbf{U}_m\in \mathds{
R}^{K\times d}$ as the output of decoder layer $m$. The coarse entity proposals are fed into the transformer decoder as the input of the first decoder layer $\mathbf{U}_0=\mathbf{Q}$. The output of each decoder layer will be fed into the next layer, forming an iterative refining process. 

\paragraph{Self-attention.} Entities in a sentence are related to each other. Therefore, modeling the relationship between different entities is helpful for NER. In self-attention layer, entity proposals interact with each other through the multi-head attention mechanism:

\begin{equation}
    \mathbf{U}^{\text{SA}}_{m}=\operatorname{MultiHeadAttn}(\mathbf{U}_{m-1},\mathbf{U}_{m-1},\mathbf{U}_{m-1})
\end{equation}

\paragraph{Cross-attention with multi-scale features.} In order to model the relationship between the proposal and other phrases in the input sentences, entity proposals interact with sentence representations through cross-attention so that richer contextual information can be aggregated into the representations of the entity proposals:

\begin{equation}
    \mathbf{U}^{\text{CA}}_{m}=\operatorname{MultiHeadAttn}(\mathbf{U}^{\text{SA}}_{m},\mathbf{H},\mathbf{H})
\end{equation}

\noindent where $\mathbf{H}$ is sentence representation. Since natural language sentences are hierarchical, we use multi-scale sentence representations to provide hierarchical contextual information for the nested NER task. Therefore, we collect the span representations generated in the propose stage to form layered pyramid-like multi-scale sentence representations:

\begin{subequations}
    \begin{equation}
        \mathbf{H}_l=[\mathbf{h}_{l,1},\mathbf{h}_{l,2},\dots,\mathbf{h}_{l,N-l+1} ]
    \end{equation}
    \begin{equation}
        \mathbf{H}=\operatorname{Flatten}([\mathbf{H}_{1},\mathbf{H}_{2},\dots,\mathbf{H}_{L} ])
    \end{equation}
\end{subequations}

\noindent where $\mathbf{H}_l\in \mathds{R}^{(N-l+1)\times d}$ is the list of features of spans with length $l$, $\mathbf{H}\in \mathds{R}^{c\times d}$ is the list of all span features, and $c=\frac{(2N-L+1)L}{2}$ is the number of the enumerated spans. Since $\mathbf{H}$ contains features of spans of different lengths, $\mathbf{H}$ can be viewed as the multi-scale representation of the input sentence. With multi-scale features, proposal representations can directly attend with features of related spans. Compared with token-level features, using multi-scale features as keys and values in cross-attention can aggregate hierarchical contextual information more effectively.

\paragraph{Feed-forward layer.} The entity proposals processed by the self-attention layer and the cross-attention layer will be fed into a feed-forward layer to generate the refined proposals of the current decoder layer:

\begin{equation}
    \mathbf{U}_{m}=\operatorname{Linear}(\operatorname{ReLU}(\operatorname{Linear}(\mathbf{U}_m^{\text{CA}})))
\end{equation}

\paragraph{Re-Prediction.} In order to eliminate the errors in coarse proposals with the information incorporated in the transformer decoder, we use the output of the last transformer decoder layer ($\mathbf{U}_{M}$) to re-predict entity classes and boundaries. For each refined proposal $\mathbf{u}_i$ in $\mathbf{U}_M$, we compute the entity classification probability of $\mathbf{u}_i$:

\begin{equation}
    \label{eq:Cls}
    \mathbf{p}^{\text{cls}}_i=\operatorname{Softmax}(\operatorname{Linear}(\mathbf{u}_i))
\end{equation}

For boundary detection, we first fuse refined entity proposal $\mathbf{u}_i$ with 1-gram span features (token-level features):

\begin{equation}
    \mathbf{H}^{\text{fuse}}_i=[[\mathbf{u}_i;\mathbf{h}_{1,1}],[\mathbf{u}_i;\mathbf{h}_{1,2}],\ldots ,[\mathbf{u}_i;\mathbf{h}_{1,N}  ]]
\end{equation}

\noindent And then we perform classification over the fused features to obtain the probability of each token to be the left and right boundary of the entity:

\begin{equation}
    \mathbf{p}^\delta_i=\operatorname{Softmax}(\operatorname{MLP}_{\delta}(\mathbf{H}^{\text{fuse}}_i)) \quad \delta \in \{\text{l},\text{r}\}
\end{equation}

\noindent where $\operatorname{MLP}$ is multi-layer perceptron.

\subsection{Training Objective}
\label{sec:Loss}

\paragraph{Proposal Loss.} We first calculate the loss of the entity proposals generated in the propose stage. The span-based entity proposal generation is a type classification task, so we use cross-entropy to obtain the loss between ground truth entity type and span-based entity classification of all enumerated spans:

\begin{equation}
    \mathcal{L}_{\operatorname{proposal}}=-\sum_{l=1}^{L}\sum_{i=1}^{N-l+1}log\mathbf{p}^{\text{cls}}_{l,i}(c_{l,i})
\end{equation}

\noindent where $c_{l,i}$ is the ground truth entity type of span $(l,i)$.

\paragraph{Refine Loss.} The final entity predictions of our PnRNet are order-agnostic, so we consider them as a set $\hat{y}=\{(\mathbf{p}_k^{\text{cls}}, \mathbf{p}_k^\text{l}, \mathbf{p}_k^\text{r})\mid k=1,\dots,K\}$. Following \citep{tan2021sequencetoset}, we compute a permutation-invariant set loss between the final entity predictions and ground truths. We first define the match cost between gold entity $y_k=(c_k,l_k,r_k)$ and prediction indexed by $\sigma(k)$:

\begin{equation}
    \begin{aligned}
        \mathcal{L}_{\operatorname{match}}(y_{k}, \hat{y}_{\sigma(k)}) &= -\mathds{1}_{\{c_{k} \neq \varnothing\}}[\\
        & \mathbf{p}_{\sigma(k)}^{\text{cls} }(c_{k})+\mathbf{p}_{\sigma(k)}^{\text{l} }(l_{k})+\mathbf{p}_{\sigma(k)}^{\text{r} }(r_{k})]
    \end{aligned}
\end{equation}

\noindent where $\mathds{1}$ denotes the indicator function. Then, we find an optimal match between prediction set and gold entities:

\begin{equation}
    \hat{\sigma}=\underset{\sigma \in \mathfrak{S}_{K}}{\arg \min } \sum_{k}^{K} \mathcal{L}_{\operatorname{match}}(y_{k}, \hat{y}_{\sigma(k)})
\end{equation}

\noindent This optimal assignment problem can be easily solved by the Hungarian algorithm \cite{kuhn1955hungarian}. The loss for the refine stage is defined as the sum of the classification loss and the boundary prediction loss of all $K$ predictions:

\begin{equation}
    \label{eq:FinalLoss}
    \begin{aligned}
        \mathcal{L}_{\text{refine}}(y, \hat{y}) &=-\sum_{k=1}^{K}\{\lambda^{\text{cls}}\log \mathbf{p}_{\hat{\sigma}(k)}^{\text{cls}}(c_{k})+\lambda^{\text{b}} \mathds{1}_{\{c_{k} \neq \varnothing\}}[\\
        &\log \mathbf{p}_{\hat{\sigma}(k)}^{\text{l}}(l_{k})+\log\mathbf{p}_{\hat{\sigma}(k)}^{\text{r}}(r_{k})]\}
    \end{aligned}
\end{equation}

\noindent where $\lambda^\text{cls}$, $\lambda^\text{b}$ are loss weights. We train the model with auxiliary losses, i.e., using the output of each decoder layer to predict entities and sum losses of all layers up for fast convergence.

\section{Experiments}

\subsection{Setting}

\paragraph{Dataset.} We conduct experiments on four wildly used nested NER datasets -- ACE04  \cite{doddington2004automatic}, ACE05 \cite{walkerchristopher2006ace}, GENIA \cite{ji2017overview}, and KBP17 \cite{ohta2002genia}. Following \cite{katiyar2018nested}, we split samples of ACE04 and ACE05 into train, dev, test set by 8:1:1, and split samples of GENIA into train/dev, test set by 9:1. For KBP17, we split all documents into 866/20/167 documents for train, dev, and test set, following \cite{lin2019sequencetonuggets}. We also conduct experiments on a flat NER dataset, CoNLL03 \cite{tjongkimsang2003introduction}. 


\paragraph{Evaluation metric.} Entity prediction is considered correct when both span and category are correctly predicted. We consider precision, recall, and F1 score as our evaluation metrics. We additionally report classification F1 score and localization F1 score in the ablation study for detailed analysis.

\begin{table}[htbp]
\centering
\small
\begin{tabular}{lccc}

\toprule
\multirow{2}{*}{Model}   & \multicolumn{3}{c}{ACE04}  \\
 \cmidrule(lr){2-4} 
& Pr.  & Rec. & F1  \\
\midrule
\citep{katiyar2018nested} &  73.60 &  71.80 &  72.70 \\
\citep{strakova2019neural}  & - & - &  84.40  \\
\citep{li2020unified} & 85.05 & 86.32 &  85.98 \\
\citep{wang2020pyramid}       & 86.08  & 86.48  & 86.28     \\
\citep{yu2020named}       & 87.30  & 86.00  & 86.70     \\
\citep{yan2021unified}  & 87.27  & 86.41  & 86.84     \\
\citep{tan2021sequencetoset}  & \textbf{88.46} &  86.10 &  87.26    \\
\citep{shen2021locate} & 87.44  & 87.38  & 87.41  \\
\midrule
PnRNet & 87.90 & \textbf{88.34} & \textbf{88.12} \\
\bottomrule
\multirow{2}{*}{Model}   & \multicolumn{3}{c}{ACE05}  \\
 \cmidrule(lr){2-4} 
& Pr.  & Rec. & F1  \\
\midrule
\citep{katiyar2018nested}  & 70.60 & 70.40 & 70.50 \\
\citep{lin2019sequencetonuggets}       & 76.20  & 73.60 & 74.90  \\
\citep{wang2020pyramid}       & 83.95  & 85.39  & 84.66     \\
\citep{yan2021unified}  & 83.16  & 86.38  & 84.74     \\
\citep{yu2020named}       & 85.20  & 85.60  & 85.40     \\
\citep{li2020unified} & 87.16 & 86.59 & 86.88 \\
\citep{shen2021locate} & 86.09   & 87.27 & 86.67  \\
\citep{tan2021sequencetoset}  & \textbf{87.48}  & 86.63  & 87.05     \\
\midrule
PnRNet    & 86.27 & \textbf{89.04} & \textbf{87.63} \\
\bottomrule
\multirow{2}{*}{Model}   & \multicolumn{3}{c}{GENIA}  \\
 \cmidrule(lr){2-4} 
& Pr.  & Rec. & F1  \\
\midrule
\citep{lin2019sequencetonuggets}       & 75.80  & 73.90 & 74.80  \\
\citep{strakova2019neural}  & - & - &  78.31 \\
\citep{wang2020pyramid}       & 79.45  & 78.94  & 79.19     \\
\citep{yan2021unified}  & 78.87  & 79.6  & 79.23     \\
\citep{tan2021sequencetoset}  & 82.31  & 78.66  & 80.44     \\
\citep{yu2020named}       & 81.80  & 79.30  & 80.50     \\
\citep{shen2021locate} & 80.19 & 80.89 & 80.54  \\
\midrule
PnRNet & \textbf{82.68} & \textbf{81.04} & \textbf{81.85} \\
\bottomrule
\multirow{2}{*}{Model}   & \multicolumn{3}{c}{KBP17}  \\
 \cmidrule(lr){2-4} 
& Pr.  & Rec. & F1  \\
\midrule
\citep{ji2017overview}       & 76.20 & 73.00 & 72.80  \\
\citep{lin2019sequencetonuggets}       & 77.70  & 71.80 & 74.60  \\
\citep{li2020unified} & 80.97  & 81.12  & 80.97  \\
\citep{tan2021sequencetoset}  & 84.91  & 83.04  & 83.96     \\
\citep{shen2021locate} & {85.46}  & {82.67}  & {84.05}  \\
\midrule
PnRNet & \textbf{86.51} & \textbf{84.06} & \textbf{85.27} \\
\bottomrule
\multirow{2}{*}{Model}   & \multicolumn{3}{c}{CoNLL03}  \\
 \cmidrule(lr){2-4} 
& Pr.  & Rec. & F1  \\
\midrule
\citep{lample2016neural}       & - & - & 90.94  \\
\citep{devlin2019bert}       & -  & - & 92.8  \\
\citep{strakova2019neural}  & - & - & 93.38 \\
\citep{wang2020pyramid}         & - & - & 93.43 \\
\citep{li2020unified} & 92.33  & 94.61 & 93.04  \\
\citep{yu2020named} & \textbf{93.7} & 93.3 & 93.5 \\
\midrule
PnRNet & 93.18 & \textbf{94.14} & \textbf{93.66} \\
\bottomrule
\end{tabular}
\caption{Main Results on four nested NER datasets (ACE04, ACE05, GENIA, KBP17) and one flat NER dataset (CoNLL03). Our PnRNet achieves state-of-the-art performance in F1-score on all these datasets. }
\label{tab:OverallPerf}
\end{table}

\begin{table*}[htb]
\centering
\small
\begin{tabular}{ccccccccccccc} 
\toprule
\multirow{2}{*}{\begin{tabular}[c]{@{}c@{}}Entity\\Proposal\end{tabular}} &
\multirow{2}{*}{\begin{tabular}[c]{@{}c@{}}Proposal\\Refinement\end{tabular}} & \multirow{2}{*}{\begin{tabular}[c]{@{}c@{}}Multi-Scale\\Features\end{tabular}} & \multicolumn{5}{c}{ACE04}                                                          & \multicolumn{5}{c}{GENIA}                                                           \\
\cmidrule(lr){4-8}  \cmidrule(lr){9-13} 
& & & Loc. F1 & Cls. F1 & Pr.  & Rec. & F1 & Loc. F1 & Cls. F1 & Pr.  & Rec. & F1  \\
\midrule
\checkmark & \checkmark & \checkmark & \textbf{92.34} & \textbf{91.75} & \textbf{87.90} & \textbf{88.34} & \textbf{88.12} & \textbf{84.65} & 88.17 & \textbf{82.68} & 81.04 & \textbf{81.85} \\
 & \checkmark & \checkmark & 89.52 & 88.40 & 81.54 & 85.40 & 83.42 & 84.13 & 87.75 & 82.58 & 80.33 & 81.44 \\
\checkmark & & & 90.35 & 91.57 & 84.40 & 88.04 & 86.18 & 84.42 & \textbf{88.87} & 81.10 & \textbf{82.20} & 81.65 \\
\checkmark & \checkmark & & 91.75 & 90.78 & 86.86 & 87.12 & 86.99 & 83.89 & 87.42 & 82.02 & 79.93 & 80.96 \\
\bottomrule
\end{tabular}
\caption{Ablation Study}
\label{tab:Ablation}
\end{table*}

\paragraph{Implementation details.} We use pre-trained BERT \cite{devlin2019bert} as the contextual encoder. For a fair comparison, we use the BERT-base-cased model for the KBP17 dataset, BERT-large-cased model for ACE04, ACE05, and CoNLL03 datasets, and BioBERT-large-cased-v1.1 \cite{lee2020biobert} for GENIA dataset. We use GloVe (100d) \cite{pennington2014glove} as our pre-trained word embedding in all experiments except GENIA and use BioWordVec \cite{chiu2016how} for the GENIA dataset. We set the span enumeration length limit to $L=16$, the number of layers of the transformer decoder to $M=3$. We pick $K=60$ proposals with the highest scores as entity proposals, a number significantly larger than the number of entities in most sentences. 

\subsection{Overall Performance}

Table \ref{tab:OverallPerf} demonstrates the overall performance of our PnRNet compared with various baselines. The experiments on nested NER datasets show that our PnRNet outperforms all previous methods by a large margin. Specifically, PnRNet achieves +0.71\%, +0.58\%, +1.31\%, and +1.22\% gain in F1-score in ACE04, ACE05, GENIA, and KBP17. On the flat NER dataset CoNLL03, PnRNet also achieves SOTA performance. It shows that modeling interactions between entities and incorporating richer hierarchical contextual information into the entity proposals not only help in detecting nested entities but also improve the performance of the flat NER.

\subsection{Ablation Study}

We conduct the ablation study in the following three aspects, as shown in Table \ref{tab:Ablation}.

\paragraph{Span-based entity proposal.} To validate the effectiveness of our proposal generation process, we replace these proposal features with a set of randomly initialized learnable embeddings. The F1-score drops by -4.7\% and -0.41\% in ACE04 and GENIA datasets without entity proposal. It shows that proposal representations generated in the first stage provide necessary information for entity recognition compared with randomly initialized vectors.

\paragraph{Proposal refinement.} In the ablation experiment without proposal refinement, we directly evaluate the performance of the span-based predictor. The performance drops by -1.94\% and -0.20\% in ACE04 and GENIA compared with full PnRNet. This indicates aggregating richer contextual information and modeling the relationship with other phrases can benefit the performance of NER.

\paragraph{Multi-scale feature.} In the ablation experiment without multi-scale features, we use the output of the sequence encoder ($\mathbf{H}^{\prime}=[\mathbf{x}_1,\mathbf{x}_2,\dots,\mathbf{x}_{N}]$), which is the token-level sentence representation, to provide contextual information in proposal refinement. The performance drops by -1.13\% and -0.89\% in ACE04 and GENIA datasets. It shows that multi-scale sentence representations provide richer hierarchical contextual information, which is helpful in entity recognition.


\subsection{Detailed Analysis of the Effect of the Proposal Refinement}

For detailed analysis, we compare the performance between span-based proposals and the final predictions of PnRNet on entities of different lengths. As shown in Table \ref{tab:DiffLen}, as the entity length grows, the performance of the span-based entity recognition (entity proposals) declines significantly. In contrast, the performance of the final prediction only gets a slight drop. Furthermore, the two-stage detection still has satisfactory performance on very long entities, even when it exceeds the prediction limits of the span-based predictor used in the propose stage. This indicates the refine stage of PnRNet, which performs interaction between proposals and incorporates multi-scale contextual information into proposal features, helps a lot in recognizing nested named entities, especially for long entities.

\begin{table}[tb]
\centering
\begin{tabular}{llcc}
\toprule
len  & support       & \multicolumn{1}{l}{Proposal} & \multicolumn{1}{l}{Final}  \\ 
\midrule
all  & 3035          & 86.18                       & \textbf{88.12}                       \\
1    & 1519 (50.0\%) & 90.49                       & \textbf{90.54}                       \\
2    & 626  (20.6\%) & 87.44                       & \textbf{88.14}                       \\
3    & 318  (10.5\%) & 85.98                       & \textbf{88.06}                       \\
4    & 149  (4.9\%)  & 83.28                       & \textbf{85.62}                       \\
5    & 107  (3.5\%)  & 80.51                       & \textbf{84.65}                       \\
6-8  & 172  (5.7\%)  & 77.01                       & \textbf{83.28}                       \\
9-16 & 105  (3.5\%)  & 62.55                       & \textbf{73.83}                       \\
17-  & 39   (1.3\%)  & -                           & \textbf{73.42}                       \\
\bottomrule
\end{tabular}
\caption{Comparison of F1-score between the entity proposals generated by the span-based predictor in the propose stage and the final prediction of PnRNet on entities of different lengths in ACE04. The span-based entity predictor cannot propose spans that exceed the span enumeration length limitation $L$, which is 16 in this experiment.}
\label{tab:DiffLen}
\end{table}

\subsection{Visualization of Multi-Scale Cross-Attention Weight}

\begin{figure}[tb]
  \centering
  \includegraphics[width=0.55\linewidth]{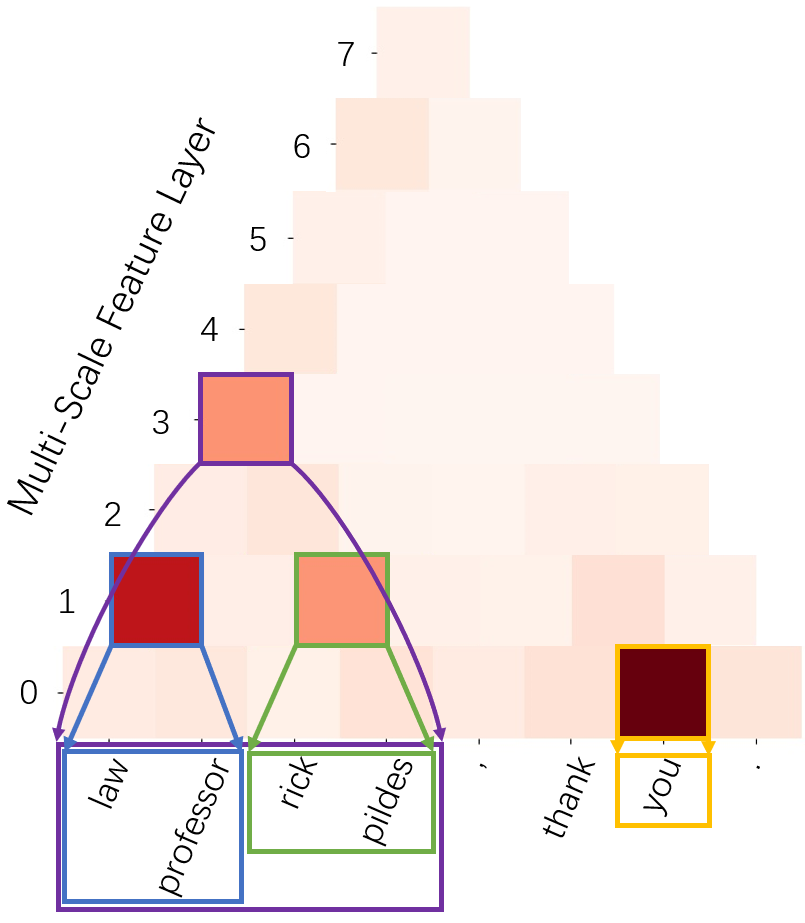}
  \caption{We show an example from the ACE04 dataset to illustrate the multi-scale attention weight of PnRNet. We visualize the cross-attention weight of a certain attention head in the last decoder layer. The query of the illustrated cross-attention weight is an entity proposal that finally predicts ``law professor"(PER). }
  \label{fig:VisAttn}
\end{figure}

We visualize the cross-attention weight map of the last decoder layer of our PnRNet to confirm the effectiveness of the multi-scale features. As shown in Figure \ref{fig:VisAttn}, four spans with the highest attention scores are ``law professor" (the predicted entity of the proposal), ``rick pildes" (the person name of the ``law professor"), ``law professor rick pildes" (an entity related to ``law professor") and "you" (another entity mentioned in this sample). This indicates that through multi-scale features, the entity proposal can directly attend to features of spans that are highly related to the proposal in cross-attention.  With the power of multi-scale features, the transformer decoder can aggregate hierarchical information that is helpful to detect named entities, improving the performance of the nested NER.


\subsection{Case Study}

We illustrate some cases in Table \ref{tab:2StageCases} to show that our PnRNet can eliminate errors in coarse proposals through proposal refinement. In case 1, by aggregating richer contextual information, boundary errors in the coarse proposal (``Arabia") can be fixed in the final prediction (``Saudi Arabia"). In case 2, the pronoun entity ``its" is misclassified by the span-based predictor as a \texttt{PER}. By interacting between the entity proposal and other proposals and contextual phrases, our PnRNet can correctly classify the entity class as an \texttt{ORG}.

\begin{table}[tb]
\small
\centering
\begin{tabular}{p{8cm}}
\toprule
`` Health officials in \textcolor{blue}{[}\textcolor{green}{[}Saudi \textcolor{red}{[}Arabia\textcolor{red}{]$_{\text{GPE}}$}\textcolor{green}{]$_{\text{GPE}}$}\textcolor{blue}{]$_{\text{GPE}}$} have asked pilgrims visiting its holy sites to wear masks in crowded places to stop the spread of the MERS coronavirus . ''\\
\midrule
On the contrary , the patent owner may continue to pursue \textcolor{red}{[}\textcolor{blue}{[}\textcolor{green}{[}its\textcolor{green}{]$_{\text{ORG}}$}\textcolor{blue}{]$_{\text{ORG}}$}\textcolor{red}{]$_{\text{PER}}$} rights through the courts .\\
\bottomrule
\end{tabular}
\caption{Case study. We mark coarse proposals in red, the corresponding final predictions in blue, and ground truths in green. For simplicity, we omit other irrelevant proposals and predictions and only show one pair of proposal and its corresponding final prediction in each case. }
\label{tab:2StageCases}
\end{table}

\section{Related Work}

Various methods have been proposed to recognize nested named entities. Since traditional sequence tagging methods \cite{huang2015bidirectional,lample2016neural} which predict a label for each token cannot address nested named entities, some optimized tagging schemes are proposed to cope with the nested NER task \cite{ju2018neural,strakova2019neural}. Hypergraph methods \cite{lu2015joint,katiyar2018nested} represent the parsing structure of the input sentence as a hypergraph and detect nested entities on the graph. Transition-based methods \cite{wang2018neural} generate a sequence of constituency parsing actions to detect nested entities.

Span-based methods predict entities with span representations. \citep{sohrab2018deep} exhaustively enumerate spans and generate span representation with boundary token features and pooling features of span tokens. \citep{tan2020boundary} first predict boundary and then perform classification over span features. \citep{wang2020pyramid} use a pyramid model to generate span representations layer by layer. \citep{yu2020named} use a bi-affine operation to compute span classification scores. \citep{shen2021locate} perform boundary regression after span-based prediction. Span-based methods can naturally address the nested NER task without complex detecting schemas and have achieved promising performance. However, span representations does not model the relationship with other contextual phrases or entities. Besides, span-based methods have difficulty predicting long entities because the span enumeration length is limited to reduce computational complexity. Our PnRNet solves all these two issues through proposal refinement and re-prediction.

Other studies design new architectures or incorporate different paradigms for the nested NER task. \citep{lin2019sequencetonuggets} first identify anchor words of entity mentions and then detect entity boundaries. \citep{li2020unified} use a machine reading comprehension model for the nested NER. \citep{yan2021unified} model the nested NER as a sequence generation task. Since the nested NER task is essentially an order-agnostic set prediction problem, \citep{tan2021sequencetoset} use a sequence-to-set neural network to detect entities as a set and apply a permutation-invariant set loss for training. However, most of these methods only use token-level encodings as sentence representations, which have difficulty representing the hierarchical structure of natural language sentences. We mitigate this issue with multi-scale sentence representation.

\section{Conclusion}

This paper presents a novel two-stage set prediction network named Propose-and-Refine Network. Firstly, we use a span-based predictor to generate a set of coarse entity predictions as proposals. Then proposals are fed into a transformer decoder for further refinement and finally re-predict entity boundaries and entity classes. So prediction errors in coarse entity proposals can be eliminated, and the model can better detect long entities. Moreover, we generate multi-scale sentence representations to provide richer hierarchical contextual information of the input sentence. Finally, we apply a cross-entropy loss for the entity proposals and a permutation-invariant set loss for the final predictions. Experiments show that our model achieves state-of-the-art performance on flat and nested NER datasets. 

\section*{Acknowledgments}

This work is supported by the National Key Research and Development Project of China (No. 2018AAA0101900), the Key Research and Development Program of Zhejiang Province, China (No. 2021C01013), CKCEST, and MOE Engineering Research Center of Digital Library.

\bibliographystyle{named}
\bibliography{ijcai22}

\appendix

\section{More Experiments}

\subsection{Span Enumeration Length Limitation}

We evaluate different span enumeration length limits on ACE04 datasets. With small span enumeration length limitations, the span-based predictor cannot propose long entities, and it is difficult for multi-scale features to model the deep hierarchical structure of natural language sentences. However, enumerating long spans is computationally expensive and may introduce noise in cross-attention. We find that $L=16$ is an optimal choice.

\begin{table}[htb]
\small
\centering
\begin{tabular}{lccccc} 
\toprule
Length & Loc. F1 & Cls. F1 & Pr & Rec. & F1  \\ 
\midrule
8     & 91.74 & 90.92 & 86.97 & 87.28 & 87.12 \\
12    & 90.93 & 90.71 & 87.83 & 86.82 & 87.32 \\
16    & \textbf{92.34} & \textbf{91.75} & \textbf{87.90} & \textbf{88.34} & \textbf{88.12} \\
20    & 92.16 & 91.30 & 87.41 & 87.81 & 87.61 \\
24    & 91.90 & 91.23 & 87.35 & 87.61 & 87.48 \\
\bottomrule
\end{tabular}
\caption{Effect of the different span enumeration length limits. Evaluate on ACE04 dataset. }
\label{tab:FpnLayerACE04}
\end{table}


\subsection{Number of Layers of the Transformer Decoder}

We try different numbers of layers of the transformer decoder on the ACE04 dataset. The transformer decoder needs multiple layers to refine the coarse proposals iteratively. We find that $M=3$ is an optimal choice.

\begin{table}[htb]
\small
\centering
\begin{tabular}{lccccc} 
\toprule
Layers & Loc. F1 & Cls. F1 & Pr & Rec. & F1  \\ 
\midrule
1    & 92.05 & 91.11 & 86.88 & 87.74 & 87.31 \\
2    & 92.01 & 91.26 & \textbf{88.01} & 87.51 & 87.76 \\
3    & \textbf{92.34} & \textbf{91.75} & 87.90 & \textbf{88.34} & \textbf{88.12} \\
4    & 91.63 & 90.73 & 87.23 & 86.89 & 87.06 \\
5    & 91.80 & 90.55 & 87.16 & 86.79 & 86.97 \\
\bottomrule
\end{tabular}
\caption{Effect of the different number of layers of the transformer decoder. Evaluate on ACE04 dataset. }
\label{tab:SSNLayerACE04}
\end{table}


\subsection{More Multi-Scale Cross-Attention Weight}

We show more examples from the ACE04 dataset to illustrate the multi-scale attention weight of PnRNet. We visualize the cross-attention weight of a certain attention head in the last decoder layer. 

\begin{figure}[htb]
  \centering
  \includegraphics[width=0.65\linewidth]{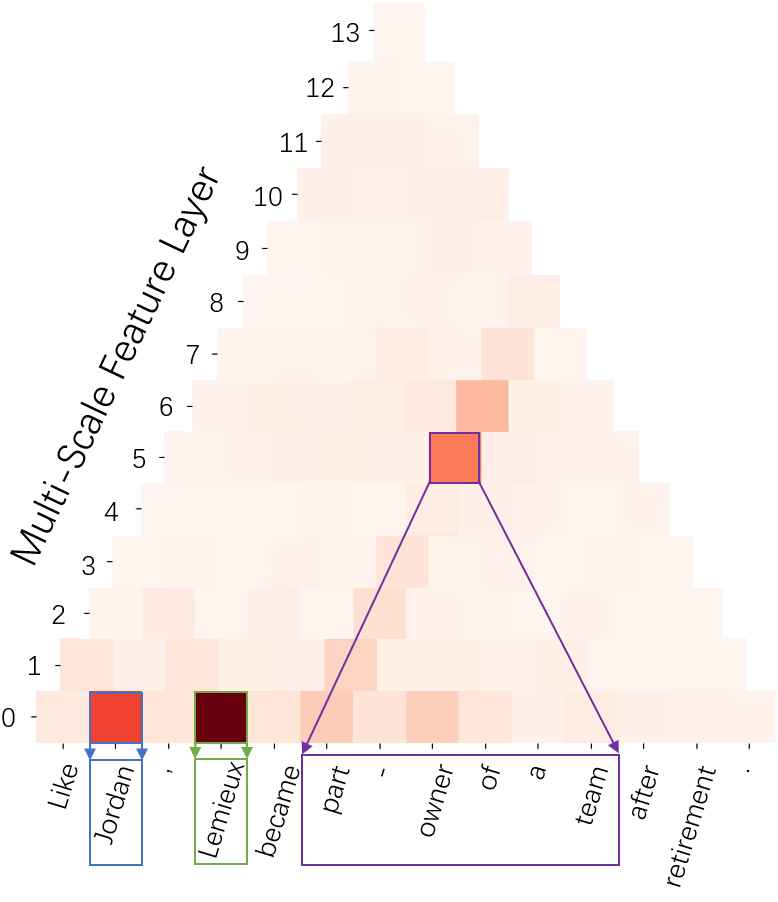}
  \caption{The query of the illustrated cross-attention weight is an entity proposal that finally predicts ``Lemieux"(PER).}
  \label{fig:MoreVis1}
\end{figure}

In Figure \ref{fig:MoreVis1}, the span with the highest score is ``Lemieux" (the predicted entity of the proposal). Another two high-score phrases are ``Jordan" (the person highly related to ``Lemieux") and ``part-owner of a team" (an entity related to ``law professor").

\begin{figure}[htb]
  \centering
  \includegraphics[width=0.65\linewidth]{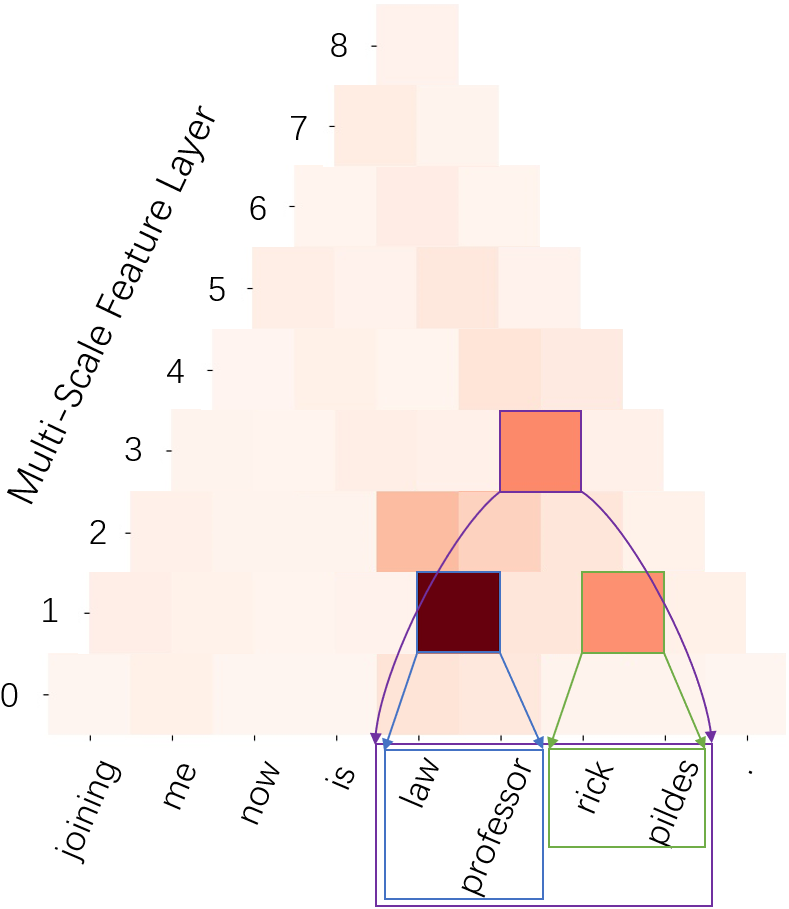}
  \caption{The query of the illustrated cross-attention weight is an entity proposal that finally predicts ``law professor rick pildes"(PER).}
  \label{fig:MoreVis2}
\end{figure}

In Figure \ref{fig:MoreVis2}, three spans with the highest scores are ``law professor rick pildes" (the predicted entity of the proposal), ``law professor", ``rick pildes" (two sub-constituents of the entity ``law professor rick pildes").

Through these cases, we can see that multi-scale features can model the hierarchical structure of natural language sentences well and can help a lot in predicting nested named entities. In our PnRNet, multi-scale features are used to provide richer hierarchical contextual information for locally contextualized entity proposals and achieve better performance compared with token-level features.


\section{Dataset Statistics}

Table \ref{tab:DsetStat} shows the statistics of the datasets we used in experiments. We show (1) the number of sentences, (2) the average sentence length, (3) the ratio of nested entities, (4) the average number of entities in a sentence, (5) the maximum number of entities in a sentence, (6) average entity length, and (7) maximum entity length.

\begin{table*}[htb]
\centering
\small
\begin{tabular}{lm{0.7cm}m{0.6cm}m{0.6cm}m{0.7cm}m{0.6cm}m{0.7cm}m{0.8cm}m{0.7cm}m{0.8cm}m{0.6cm}m{0.7cm}m{0.8cm}m{0.7cm}m{0.7cm}} 
\toprule \multirow{2}{*}{\begin{tabular}[c]{@{}c@{}}Statistics\end{tabular}} & \multicolumn{3}{c}{ACE04}                                                          &  \multicolumn{3}{c}{ACE05}                                                          & \multicolumn{2}{c}{GENIA}                                                          & \multicolumn{3}{c}{KBP17}                                                          &
\multicolumn{3}{c}{CoNLL03}                                                           \\
\cmidrule(lr){2-4}  \cmidrule(lr){5-7}   \cmidrule(lr){8-9}   \cmidrule(lr){10-12}   \cmidrule(lr){13-15} 
& train & dev & test & train & dev & test & train & test & train & dev & test & train & dev & test \\
\midrule
\#sent         & 6200 & 745  & 812  & 7194 & 969  & 1047 & 16692 & 1854 & 10546 & 545  & 4267 & 10773 & 3250 & 3453  \\
avg sent len   & 22.5 & 23.0 & 23.1 & 19.2 & 18.9 & 17.2 & 25.3  & 26.0 & 19.6  & 20.6 & 19.3 & 15.9  & 15.8 & 13.4  \\
nested ratio   & 45.7 & 43.4 & 46.7 & 38.4 & 34.8 & 37.4 & 17.9  & 21.8 & 28.1  & 32.2 & 29.4 & -     & -    & -     \\
avg \#e / sent & 3.6  & 3.4  & 3.7  & 3.4  & 3.3  & 2.9  & 3.0   & 3.0  & 3.0   & 3.4  & 3.0  & 2.1   & 1.8  & 1.6   \\
max \#e / sent & 28   & 22   & 20   & 27   & 23   & 17   & 25    & 14   & 58    & 15   & 21   & 20    & 20   & 31    \\
avg e-len      & 2.6  & 2.7  & 2.7  & 2.3  & 2.1  & 2.3  & 2.0   & 2.1  & 1.9   & 2.1  & 2.0  & 1.4   & 1.4  & 1.4   \\
max e-len      & 63   & 40   & 44   & 49   & 31   & 27   & 18    & 15   & 45    & 32   & 49   & 10    & 10   & 6    \\
\bottomrule
\end{tabular}
\caption{Statistics of datasets we used in experiments.}
\label{tab:DsetStat}
\end{table*}

\end{document}